\documentclass[letterpaper]{article} 
\usepackage[submission]{aaai24}  
\usepackage{times}  
\usepackage{helvet}  
\usepackage{courier}  
\usepackage[hyphens]{url}  
\usepackage{graphicx} 
\usepackage{natbib}  
\usepackage{caption} 
\usepackage{amsmath}
\usepackage{threeparttable}
\usepackage{color}
\usepackage{subfig}
\usepackage{multirow}
\usepackage{cite}
\usepackage{soul}
\usepackage{amssymb}
\usepackage{makecell} 

\usepackage[export]{adjustbox} 

\newtheorem{myPro}{Problem}
\newtheorem{myDef}{Definition}

\usepackage{algorithm}
\usepackage{algorithmic}

\usepackage{newfloat}
\usepackage{listings}
\DeclareCaptionStyle{ruled}{labelfont=normalfont,labelsep=colon,strut=off} 
\floatstyle{ruled}
\newfloat{listing}{tb}{lst}{}
\floatname{listing}{Listing}
\title{Urban Region Embedding via Multi-View Contrastive Prediction}

\author{
    Zechen Li\textsuperscript{\rm 1},
    Weiming Huang\textsuperscript{\rm 2},
    Kai Zhao\textsuperscript{\rm 3},
    Min Yang\textsuperscript{\rm 1},
    Yongshun Gong\textsuperscript{\rm 1},
    Meng Chen\textsuperscript{\rm 1}\thanks{Corresponding author.}
}
\affiliations{
    \textsuperscript{\rm 1} School of Software, Shandong University\\
    \textsuperscript{\rm 2} School of Computer Science and Engineering, Nanyang Technological University\\
    \textsuperscript{\rm 3} Robinson College of Business, Georgia State University\\

     lizechenn@gmail.com, weiming.huang@ntu.edu.sg, kzhao4@gsu.edu, myang3@sdu.edu.cn, yongshun2512@hotmail.com, mchen@sdu.edu.cn
}

\usepackage{bibentry}

\begin{document}
\maketitle

\begin{abstract}

Recently, learning urban region representations utilizing multi-modal data (information views) has become increasingly popular, for deep understanding of the distributions of various socioeconomic features in cities. However, previous methods usually blend multi-view information in a posteriors stage, falling short in learning coherent and consistent representations across different views. In this paper, we form a new pipeline to learn consistent representations across varying views, and propose the multi-view Contrastive Prediction model for urban Region embedding (ReCP), which leverages the multiple information views from point-of-interest (POI) and human mobility data. Specifically, ReCP comprises two major modules, namely an intra-view learning module utilizing contrastive learning and feature reconstruction to capture the unique information from each single view, and inter-view learning module that perceives the consistency between the two views using a contrastive prediction learning scheme. We conduct thorough experiments on two downstream tasks to assess the proposed model, i.e., land use clustering and region popularity prediction. The experimental results demonstrate that our model outperforms state-of-the-art baseline methods significantly in urban region representation learning.



\end{abstract}

\section{Introduction}
A deep understanding of the spatial distribution of various socioeconomic factors in cities such as land use or population distribution, is important for urban planning and management. In recent years, an increasingly popular trend in the community of urban computing has been to partition a city into numerous regions and utilize various urban sensory data to learn the latent representations of the regions, which can subsequently be used in varying urban sensing tasks, e.g., land usage clustering. house price prediction, and population density inference \cite{liu2021discovering,li2022predicting,liu2023knowledge, huang2023learning, xu2023spatial, li2023urban}. This trend can also be attributed to the prosperity of mobile sensing technologies, which has led to the rapid accumulation of urban sensing data, such as human trajectories or points-of-interest (POIs) \cite{chen2018pcnn, zhang2022beyond, xu2023tme, zhang2023towards}. Such various urban data provide more opportunities for tackling the problem of region representation learning.


\begin{figure}[!t]
    \centering
    
  \subfloat{
    \includegraphics[width=0.22\textwidth]{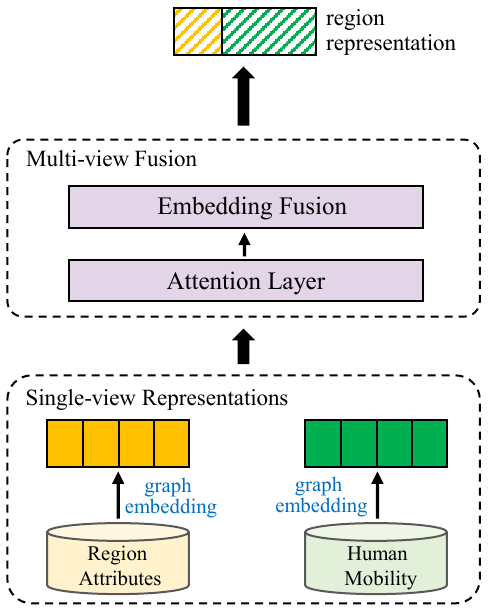}
    \hspace{1mm}
 
    \includegraphics[width=0.22\textwidth]{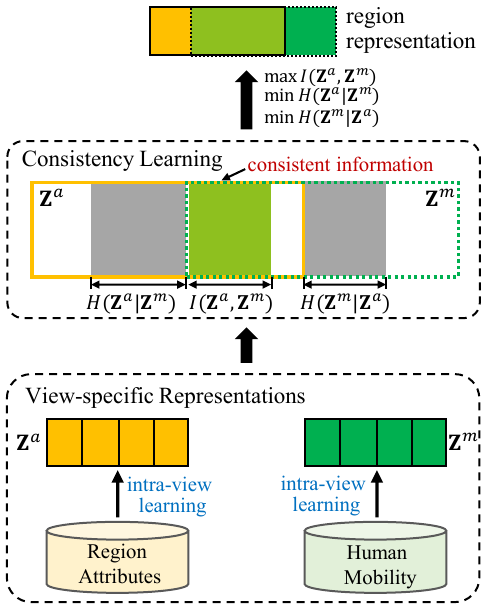}}
    \caption{Illustration of (a) multi-view fusion paradigm and our proposed (b) consistency learning paradigm for region embedding. In the right figure, the solid and dotted rectangles denote the region representations $\mathbf{Z}^a$ and $\mathbf{Z}^m$ from the attribute and mobility views, respectively. The mutual information $I(\mathbf{Z}^a, \mathbf{Z}^m)$ (chartreuse area) quantifies the amount of information shared by $\mathbf{Z}^a$ and $\mathbf{Z}^m$; the conditional entropy $H(\mathbf{Z}^a|\mathbf{Z}^m)$ (grey area) quantifies the amount of information of $\mathbf{Z}^a$ conditioned on $\mathbf{Z}^m$. To learn consistent region representations across different views, it is encouraged to maximize $I(\mathbf{Z}^a, \mathbf{Z}^m)$ and minimize $H(\mathbf{Z}^a|\mathbf{Z}^m)$ and $H(\mathbf{Z}^m|\mathbf{Z}^a)$.}
    \label{fig:intro}
\end{figure}

Many previous studies have attempted to learn region representations by utilizing human mobility data. For instance, Wang et al. \cite{wang2017region} construct flow graphs and spatial graphs using taxi flow data and propose a graph embedding method to learn region representations. Yao et al. \cite{yao2018representing} extract human mobility patterns from taxi trajectories, and model the co-occurrence of origin-destination regions to learn region representations. The above methods merely rely on single-view data, which offers a limited perspective of regions and fails to provide a comprehensive representation. Further, recent studies \cite{zhang2021multi,luo2022urban,zhou2023heterogeneous} propose learning region representations through integrating data in multiple modalities, thus forming multiple information views. In this context, the technical focus of recent region embedding studies has shifted towards the fusion between multiple information views, where they usually follow the same pipeline: separate single-view representation followed by multiple-view fusion. Such a pipeline is demonstrated in Figure~\ref{fig:intro}(a), where, it (1) separately models each information view (usually with a graph structure) and learns multiple single-view representations for each region, and (2) leverages certain fusion techniques (e.g., based on attention mechanisms) to blend multiple representations and yield the final multi-view region representation.

The previous multi-view region embedding methods have been effective in certain analyses, but they come with a notable shortcoming: neglecting the information consistency across different views when generating the final region representation. Intuitively, the information carried by multiple views of a region is highly correlated, and thus their representations should be consistent. For example, an entertainment region could contain multiple bars and restaurants (region attribute view based on POIs), as well as a large number of nighttime mobility flows (human mobility view).  Both views can reflect the intrinsic characteristics of this region (i.e., entertainment function). If we manage to leverage such correlation, it could be served as the constraint during the process of learning representations for each view, and enable the knowledge of transferring from one view to the other. Ultimately, the multi-view representations would become highly consistent and naturally fused.



Following the ideas above, we present a new pipeline - consistency learning paradigm - for multi-view region embedding from an information theory perspective \cite{tsai2021self, lin2021completer}, where the multi-view representations are naturally fused through exchanging information between views along with learning view-specific region representations, rather than treating fusion as a posterior process. This new pipeline is shown in Figure~\ref{fig:intro}(b). Given two view-specific region representations $\mathbf{Z}^a$ and $\mathbf{Z}^m$ (where they are from the region attribute view and the human mobility view, respectively), we maximize the mutual information $I(\mathbf{Z}^a, \mathbf{Z}^m)$ to increase the amount of the shared information (consistency) in the region representations of the two views. We also minimize the conditional entropy $H(\mathbf{Z}^a|\mathbf{Z}^m)$ and $H(\mathbf{Z}^m|\mathbf{Z}^a)$ to diminish the inconsistent information across the two views and improve the consistency further.


Based on the consistency learning paradigm, we propose a multi-view Contrastive Prediction model for urban Region embedding (ReCP), which can effectively enhance the consistency of region representations across different views. ReCP consists of two major components: intra-view learning and inter-view learning. In the intra-view learning component, to learn view-specific region representations, we compare each region with other dissimilar ones to embed the region into a latent space via contrastive learning; in the meantime, we also utilize autoencoders to capture view-specific region features for different views, which helps avoid model falling into a trivial solution. In the inter-view learning component, to learn the cross-view consistency of region representations, we design inter-view contrastive learning by maximizing $I(\mathbf{Z}^a, \mathbf{Z}^m)$ and dual prediction between views by minimizing $H(\mathbf{Z}^a|\mathbf{Z}^m)$ and $H(\mathbf{Z}^m|\mathbf{Z}^a)$.

To summarize, our contributions are as follows:
\begin{itemize}
\item We form a new pipeline following a consistency learning paradigm, to study the urban region embedding problem by exploring the consistency across different views, using both human mobility and POI data. Different from existing multi-view region embedding methods which adopt the attention mechanisms to fuse representations of different views, we propose to learn consistent multi-view representations of regions by increasing the amount of shared information across multiple views from the information entropy perspective.

\item  We design the inter-view contrastive learning and dual prediction processes to diminish the inconsistent information across views and learn an informative and consistent region representation between different views, achieved by maximizing the mutual information among different views and minimizing the conditional entropy among them.

\item We conduct extensive experiments to evaluate our model with real-world datasets. The results demonstrate that the proposed ReCP outperforms existing methods on two downstream tasks by a margin. Data and source code are available at https://anonymous.4open.science/r/ReCP.
\end{itemize}

\section{Problem Formulation}
We formulate the urban region representation learning problem with the following definitions:

\begin{myDef}[Urban Region] A city can be partitioned into $n$ disjoint urban regions, denoted as $\mathcal{R} =\left\{ r_1,r_2,...,r_n \right\}$.
\end{myDef}


\begin{myDef}[Region Attributes] In this study, region attributes are defined as inherent geographic features of regions. Specifically, we consider Point of Interest (POI) categories as region attributes following \cite{zhang2022region, fu2019efficient}. These region attributes are represented as a set $\mathcal{A} = \{\mathbf{A}_1, \mathbf{A}_2, \cdots, \mathbf{A}_n\}$, where $\mathbf{A}_i \in \mathbb{R}^{F}$ and $F$ represents the total number of POI categories. Each dimension in $\mathbf{A}_i$ corresponds to the number of POIs with a specific category in the region $r_i$.
\end{myDef}


\begin{myDef}[Human Mobility] For a region $r_i$, we define its outflow feature $\mathbf{S}_{i}^{j,t}$ as the number of trips made by all individuals originating from region $r_i$ and destined for region $r_j$ during a specific time interval $t$. Consequently, we generate a collection of outflow features based on the mobility data encompassing all regions within the set $\mathcal{R}$. This collection is represented as $\mathcal{S} = \{\mathbf{S}_1, \mathbf{S}_2, \cdots, \mathbf{S}_n\}$, where $\mathbf{S}_i \in \mathbb{R}^{M}$. Here, $M$ is calculated as the product of the number of regions, $n$, and the number of time intervals, $N_t$, within a day, for instance, 24. Similarly, by considering $r_i$ as the destination region and the other regions $r_j$ as the source regions, we can obtain an inflow feature vector, denoted as $\mathbf{D}_i$, and finally obtain a collection $\mathcal{D} = \{\mathbf{D}_1, \mathbf{D}_2, \cdots, \mathbf{D}_n\}$ of inflow features for all regions.
\end{myDef}


\begin{myPro}[Region Representation Learning] Given the attribute features $\mathcal{A}$, outflow features $\mathcal{S}$, and inflow features $\mathcal{D}$ of $n$ regions, our objective is to acquire a collection of low-dimensional embeddings $\mathcal{E} = \{ \mathbf{E}_1, \mathbf{E}_2, \cdots,  \mathbf{E}_n\}$, to serve as the latent representation for each region. 
\end{myPro}

\begin{figure*}[!t]
  \centering
  \includegraphics[width=0.85\textwidth]{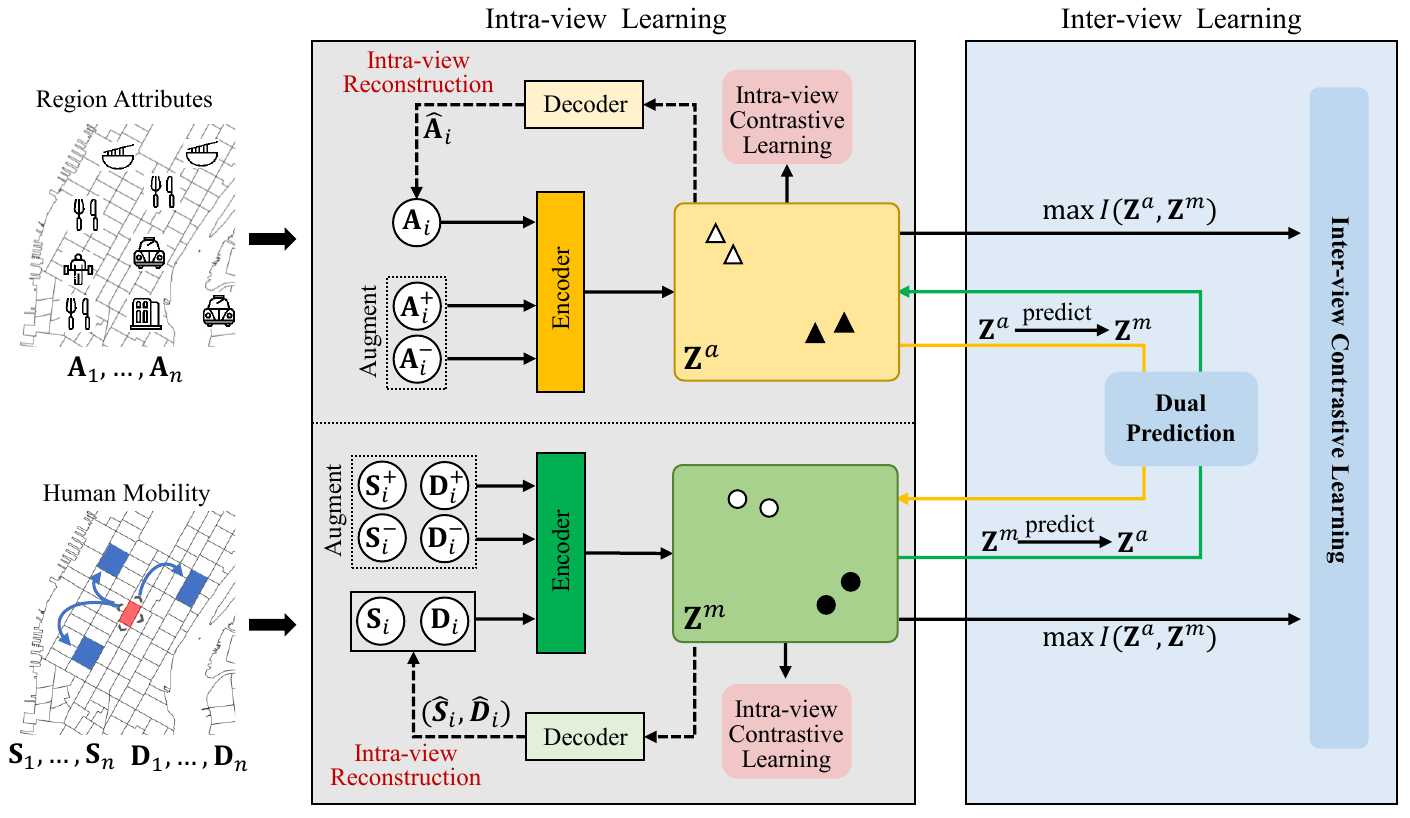}
  \caption{The framework of ReCP.}
  \label{fig:framework}
\end{figure*}

\section{Methodology}
The framework of ReCP is illustrated in Figure \ref{fig:framework}, which includes two major components:
1) intra-view learning: for both region attribute and human mobility view, it captures the representative features of each region by intra-view contrastive learning to learn view-specific representations. Additionally, feature reconstruction is designed within each view to recover the original feature of the region, which helps avoid a trivial solution; 
2) inter-view learning: within the same region, it integrates representations from different views through two learning objectives: inter-view contrastive learning is used to enhance the consistency across different views, and dual prediction is introduced to further diminish the inconsistent information between views.

\subsection{Intra-view Learning}
Initially, we learn view-specific region representations based on the region attribute features $\mathcal{A}$ and the mobility features $\mathcal{S}$ and $\mathcal{D}$, respectively. 
Within each view, we learn the latent representation for each region by employing intra-view contrastive learning, i.e., we compare each region with others to highlight distinctive features within each region. Additionally, we design a within-view reconstruction loss to avoid the trivial solution.

\subsubsection{Intra-view Contrastive Learning}
To learn region representations within each view, we design an intra-view contrastive learning module, which compares each region with others.
For a given region $r_i$, we have three types of region features, including the attribute feature $\mathbf{A}_i$, outflow feature $\mathbf{S}_i$, and inflow feature $\mathbf{D}_i$. For simplicity, let $\mathbf{X}_{i}^{v} $ denote the raw feature for the $v$-th view. For a target region $r_i$, we define its positive set as $\mathcal{P} ^v_i=\left\{ \mathbf{X}^v_{1},\mathbf{X}^v_{2}, \cdots, \mathbf{X}^v_{K}\right\}$, where $\mathbf{X}^v_{1},\mathbf{X}^v_{2}, \cdots, \mathbf{X}^v_{K}$ are positive samples obtained through the data augmentation function following \cite{zhang2022region}, and $K$ is the number of positive samples. The negative set $\mathcal{N}^{v}_i$ is defined as $\mathcal{N}^v_i = \{\mathbf{X}^v_t|t \ne  i\}$, which contains features of regions except $r_i$. 

We then map the raw features of regions into a latent representation,
\begin{equation}
\label{eq1}
\mathbf{Z}_{i}^{v}=E^{\left( v \right)}(\mathbf{X}^v_i),
\end{equation}
where $E^{(v)}$ denotes the encoder for the $v$-th view. In practice, we simply implement it as a fully connected neural network. As a result, we obtain three types of region representations, $\mathbf{Z}^{a}_i$, $\mathbf{Z}^{s}_i$ and $\mathbf{Z}^{d}_i$. Further, we compute the region representation $\mathbf{Z}^{m}_i$ of the human mobility view as the average of $\mathbf{Z}^{s}_i$ and $\mathbf{Z}^{d}_i$, i.e., $\mathbf{Z}^{m}_i=\left( \mathbf{Z}^{s}_i+\mathbf{Z}^{d}_i \right)/2$. 
To maximize the similarity of positive pairs while minimizing the similarity of negative pairs, the contrastive learning loss for the $v$-th view is defined as follows, 
\begin{equation}
\begin{aligned}
\mathcal{L} _{cl}^{v}= &\sum_{r_i\in \mathcal{R}} [-\log \sum_{k=1}^K {\exp(\frac{\mathbf{Z}_i^v \cdot \mathbf{Z}_k^v}{\tau})} + \\
& \log ( \sum_{k=1}^K {\exp(\frac{\mathbf{Z}_{i}^{v} \cdot \mathbf{Z}_{k}^{v}}{\tau})}+\sum_{t=1}^{|\mathcal{N}^v_i|} {\exp(\frac{\mathbf{Z}_i^v \cdot \mathbf{Z}_t^v}{\tau})}) ],
\end{aligned}
\end{equation}
where $\tau$ is the temperature parameter and $\mathcal{R}$ is the set of regions. Further, the intra-view contrastive learning loss across all views is formulated as
\begin{equation}
\label{eq2}
\mathcal{L}_{cl}^{intra}= \mu\mathcal{L}_{cl}^a+  \mathcal{L}_{cl}^m .
\end{equation}
where $\mu$ is the parameter controlling the balance between the attribute view and the mobility view.

\subsubsection{Intra-view Reconstruction}
Given the feature $\mathbf{X}_{i}^{v}$ for the $v$-th view of the region $r_i$, we further optimize the latent region representations via an autoencoder and define the reconstruction loss $\mathcal{L} _{rec}^v$ as
\begin{equation}
\label{reconstruction}
\mathcal{L} _{rec}^v=\sum_{r_i\in \mathcal{R}}{\left\| \mathbf{X}^v_i-D^{\left( v \right)}(E^{\left( v \right)}(\mathbf{X}^v_i)) \right\|_2^2},
\end{equation}
where $E^{(v)}$ is the same as that in Equation~(\ref{eq1}) and $D^{(v)}$ is the decoder for the $v$-th view to reconstruct the region features. Specifically, we employ a fully connected network to implement $D^{(v)}$, which shares the same number of layers and hidden sizes as $E^{(v)}$. Note that the autoencoder structure is
helpful to avoid the trivial solution. 

The total reconstruction loss across all views is 
\begin{equation}
\mathcal{L}^{intra}_{rec}= \mu\mathcal{L}_{rec}^a+  \mathcal{L}_{rec}^m,
\end{equation}
where $\mu$ is the same weight parameter as that in Equation~(\ref{eq2}). So far, we obtain two types of view-specific region representations $\mathbf{Z}_{i}^{a}$ and $\mathbf{Z}_{i}^{m}$ from the region attribute and human mobility views. 

\subsection{Inter-view Learning} \label{dms}
Different views of a region provide valuable information for describing the region, often offering complementary insights.
To learn consistent and informative representations across different views, we  employ inter-view contrastive learning to improve collaboration and information exchange between the views, achieved by maximizing the mutual information among different views. Additionally, dual prediction between two views is leveraged to reduce the impact of inconsistent information between the views by minimizing the conditional entropy across them.

\subsubsection{Inter-view Contrastive Learning}

In the latent embedding space, we conduct contrastive learning to learn consistent representations shared across different views, as recent contrastive learning studies \cite{he2020momentum, lin2021completer} have shown that consistency could be learned by maximizing the mutual information of different views.  
Formally, given the two representations $\mathbf{Z}_{i}^{a}$ and $\mathbf{Z}_{i}^{m}$ of region $r_i$, we maximize the mutual information between $\mathbf{Z}_{i}^{a}$ and $\mathbf{Z}_{i}^{m}$ from different views: 
\begin{equation}\label{equa:intra_cl}
\mathcal{L}^{inter}_{cl}=-\sum_{r_i\in \mathcal{R}} [I(\mathbf{Z}_{i}^{a}, \mathbf{Z}_{i}^{m}) +\alpha ( H( \mathbf{Z}_{i}^{a} ) +H( \mathbf{Z}_{i}^{m}))],
\end{equation}
where $I(\cdot)$ represents mutual information, $H(\cdot)$ denotes information entropy, and the parameter $\alpha$ controls the balance between mutual information and information entropy. Note that the maximization of $H\left( \mathbf{Z}_{i}^{a} \right)$ and $H\left( \mathbf{Z}_{i}^{m} \right)$ also helps prevent trivial solutions where all regions are represented by the same representation.
Based on the definition of mutual information, $I(\cdot)$ is defined as 
\begin{equation}
I\left( \mathbf{Z}^{a}_i,\mathbf{Z}^{m}_i \right)=  P \left( \mathbf{Z}^{a}_i,\mathbf{Z}^{m}_i \right) log\frac{P \left( \mathbf{Z}^{a}_i,\mathbf{Z}^{m}_i \right) }{P \left( \mathbf{Z}^{a}_i \right) P \left( \mathbf{Z}^{m}_i \right)},
\end{equation}
where $P \left( \mathbf{Z}^{a}_i, \mathbf{Z}^{m}_i \right) $ represents the joint probability distribution of $\mathbf{Z}^{a}_i$ and $\mathbf{Z}^{m}_i$. 
To represent the joint probability distribution, we employ a softmax function to transform the region representations $\mathbf{Z}_{i}^{a} \in \mathbb{R}^d$ and $\mathbf{Z}_{i}^{m} \in \mathbb{R}^d$ (where $d$ is the dimension of region representations) with
\begin{equation}
\mathbf{B}_{i}^{a}=\mathrm{softmax} \left( \mathbf{Z}_{i}^{a} \right) ,\mathbf{B}_{i}^{m}=\mathrm{softmax} \left( \mathbf{Z}_{i}^{m} \right),
\end{equation}
where $\mathbf{B}_{i}^{a} \in \mathbb{R}^d$ and $\mathbf{B}_{i}^{m} \in \mathbb{R}^d$ can be interpreted as the probability distributions. 
Considering the entire set $\mathcal{R}$ containing $n$ regions, we define the matrix $\mathbf{M} \in \mathbb{R}^{d \times d}$ as the joint probability distribution of $\mathbf{Z}^a$ and $\mathbf{Z}^m$,
\begin{equation}
\mathbf{M}=\frac{1}{n}\sum^n_{i=1}{\mathbf{B}_{i}^{a}\left( \mathbf{B}_{i}^{m} \right)}^{\mathrm{T}}.
\end{equation}
We denote the element located at the $r$-th row and the $r\prime$-th column of the matrix as $\mathbf{M}_{rr\prime}$, and the sum of the elements in matrix $\mathbf{M}$ along the $r$-th row (the $r\prime$-th column) as $\mathbf{M}_{r}$ ($\mathbf{M}_{r\prime}$). $\mathbf{M}_{rr\prime}$ represents the joint probability, while $\mathbf{M}_{r}$ and $\mathbf{M}_{r\prime}$ represent the marginal probability, respectively. 
Then we could compute the mutual information $I\left( \mathbf{Z}^{a},\mathbf{Z}^{m} \right)$ as follows,
\begin{equation}
\label{mutual}
I\left( \mathbf{Z}^{a},\mathbf{Z}^{m} \right)=\sum^d_{r=1}{\sum^d_{r\prime=1}{\mathbf{M}_{rr\prime}\log \frac{\mathbf{M}_{rr\prime}}{\mathbf{M}_{r}\cdot \mathbf{M}_{r\prime}}}}.
\end{equation}

Information entropy $H(\mathbf{Z}^{v}_i)$ is defined as follows, 
\begin{equation}
H(\mathbf{Z}^{v}_i)= - P(\mathbf{Z}^{v}_i) log{P(\mathbf{Z}^{v}_i) },
\end{equation}
where $v\in \{a,m\}$.
Following the above definition of $\mathbf{M}$, $H(\mathbf{Z}^{v}_i)$ could be computed as
\begin{equation}
\label{entropy}
\begin{aligned}
 H( \mathbf{Z}^{a} )&=-\sum^d_{r=1}{\mathbf{M}_{r}\log {\mathbf{M}_{r}}},\\
H( \mathbf{Z}^{m} )&=-\sum^d_{r\prime=1}{\mathbf{M}_{r\prime}\log {\mathbf{M}_{r\prime}}}.   
\end{aligned}
\end{equation}

Combining Equations~(\ref{equa:intra_cl}), (\ref{mutual}), and (\ref{entropy}), the inter-view contrastive learning loss is formulated as
\begin{equation}
\mathcal{L}^{inter}_{cl}=-\sum^d_{r=1}{\sum^d_{r\prime=1}{\mathbf{M}_{rr\prime}\ln \frac{\mathbf{M}_{rr\prime}}{\mathbf{M}_{r}^{\alpha +1}\cdot \mathbf{M}_{r\prime}^{\alpha +1}}}}.
\end{equation}
where $\alpha$ is the weight parameter defined in the Equation~(\ref{equa:intra_cl}).

\subsubsection{Inter-view Dual Prediction}

To further diminish the inconsistency across different views, we predict the view-specific region representation by minimizing the conditioned entropy. Formally, given the region representations $\mathbf{Z}^{a}$ and $\mathbf{Z}^{m}$, we minimize the conditional entropy $H(\mathbf{Z}^p|\mathbf{Z}^q)$, where $p=a, q=m$ or $p=m, q=a$. On one hand, $\mathbf{Z}^{q}$ contains nearly all the information required to represent the $p$-th view if $\mathbf{Z}^{q}$ can perfectly predict $\mathbf{Z}^{p}$ for any $(\mathbf{Z}^{p},\mathbf{Z}^{q}) \sim  {P}_{\mathbf{Z}^p,\mathbf{Z}^q}$. 
On the other hand, $\mathbf{Z}^{q}$ diminishes the inconsistent information within the $q$-th view if $\mathbf{Z}^{p}$ can perfectly predict $\mathbf{Z}^{q}$ under the constraint where $I(\mathbf{Z}^{p}, \mathbf{Z}^{q})$ is maximized. Mathematically, $H(\mathbf{Z}^p | \mathbf{Z}^q) $ is defined as
\begin{equation}
H\left( \mathbf{Z}^p\left| \mathbf{Z}^q \right. \right) = -{\mathbb{E} _{{P}}}_{_{\mathbf{Z}^p,\mathbf{Z}^q}}\left[ \log {P} \left( \mathbf{Z}^p|\mathbf{Z}^q \right) \right] .
\end{equation}
To minimize $H\left( \mathbf{Z}^p\left| \mathbf{Z}^q \right. \right)$, a common approach is to assume a variational distribution ${Q} \left( \mathbf{Z}^p|\mathbf{Z}^q \right)$ for $\mathbf{Z}^{p}$ and $\mathbf{Z}^{q}$. Specially, we present to maximize ${\mathbb{E} _{{P}}}_{_{\mathbf{Z}^p,\mathbf{Z}^q}}\left[ \log {Q} \left( \mathbf{Z}^p|\mathbf{Z}^q \right) \right] $, which serves as a lower bound of  ${\mathbb{E} _{{P}}}_{_{\mathbf{Z}^p,\mathbf{Z}^q}}\left[ \log {P} \left( \mathbf{Z}^p|\mathbf{Z}^q \right) \right] $. ${Q} \left( \cdot |\cdot \right)$ can be any distribution such as Gaussian or Laplacian. In this work, we simply adopt the Gaussian distribution ${N} \left. (\mathbf{Z}^p|F^{\left( q \right)}\left( \mathbf{Z}^q \right) ,\sigma \mathbf{I} \right. )$, where $F^{\left( q \right)}\left( \cdot \right) $ represents a parameterized function mapping $\mathbf{Z}^q$ to $\mathbf{Z}^p$, and $\sigma \mathbf{I}$ denotes the variance matrix. 
By ignoring the constants derived from the Gaussian distribution, maximizing ${\mathbb{E} _{{P}}}_{_{\mathbf{Z}^p,\mathbf{Z}^q}}\left[ \log {Q} \left( \mathbf{Z}^p|\mathbf{Z}^q \right) \right] $ is equivalent to minimizing
\begin{equation}
 {\mathbb{E} _{{P}}}_{_{\mathbf{Z}^p,\mathbf{Z}^q}}\left\| \mathbf{Z}^p-F^{(q)}\left( \mathbf{Z}^q \right) \right\| ^2_2.
\end{equation}

Then the dual prediction loss can be formulated as
\begin{equation}
\nonumber
\mathcal{L} _{dp}^{inter}=
\sum_{r_i\in \mathcal{R}}{ \left\| \mathbf{Z}_{i}^{m}-F^{\left( a \right)}\left( \mathbf{Z}_{i}^{a} \right) \right\| ^2_2 
+\left\| \mathbf{Z}_{i}^{a}-F^{\left( m \right)}\left( \mathbf{Z}_{i}^{m} \right) \right\| ^2_2  }.
\end{equation}

Here, $F^{\left( a \right)} $ and $F^{\left( m \right)} $ are respectively implemented as fully-connected networks, with each layer followed by a batch normalization layer and a ReLU layer. Note that the above loss may lead to model collapse without the intra-view reconstruction loss (Equation~(\ref{reconstruction})), i.e., $\mathbf{Z}_{i}^{a}$ and $\mathbf{Z}_{i}^{m}$ from different views become equivalent to the same constant. 

Finally, the inter-view learning loss is defined as
\begin{equation}
\mathcal{L} _{inter}=\mathcal{L} _{dp}^{inter}+\mathcal{L} _{cl}^{inter}.
\end{equation}

\subsection{Model Training}
The final objective function is defined as 
\begin{equation}
\mathcal{L}=\mathcal{L}_{inter}+\lambda_1\mathcal{L}^{intra}_{cl}+\lambda_2\mathcal{L}^{intra}_{rec},
\end{equation}
where $\lambda_1$ and $\lambda_2$ are parameters controlling the weights of different losses. 
After learning the latent representations $\mathbf{Z}^{a}$ and $\mathbf{Z}^{m}$, we simply concatenate them as the final multi-view region representation, i.e., $\mathbf{E}_i= \mathbf{Z}^{a}_i\left |  \right |  \mathbf{Z}^{m}_i$.

\section{Experiments}\label{experiment}
We start by presenting the experimental settings, followed by the evaluation of the learned region representations on two popular downstream tasks: land use clustering and region popularity prediction. 

\subsection{Experimental Settings}
\textbf{Datasets}.  We collect a diverse set of real-world data from NYC Open Data\footnote{https://opendata.cityofnewyork.us} and use the Manhattan borough as the study area. We partition Manhattan into 270 regions based on the city boundaries designed by the US Census Bureau\footnote{https://www.census.gov/data.html}. 
As for the human mobility data, we employ complete taxi trip records from February 2014 as our training data. We utilize the NYC check-in and POI data provided by \cite{yang2014modeling} for our model training and the popularity prediction task. The detailed description of datasets is shown in Table~\ref{tab:Dataset}. Based on these data, we construct the region features including $\mathcal{A}$, $\mathcal{S}$, and $\mathcal{D}$ for model training.

\begin{table}[!t]
\centering
\caption{Data description (K=$10^3$, M=$10^6$).}\label{tab:Dataset}
\begin{tabular}
{c|c}
\hline
\textbf{Dataset} & \textbf{Description} \\

\hline
 \thead{Regions} & \thead{270 regions divided by streets in Manhattan}  \\
 \hline
  \thead{Taxi trips} & \thead{10M taxi trips during February, 2014}  \\
  \hline
   \thead{ POI data} & \thead{ 10K POIs with 244 categories}  \\
   \hline
   \thead{ Check-in data} & \thead{ 100K check-in records}  \\
\hline

\end{tabular}
\end{table}

\begin{table*}[!th]
\centering
\caption{Performance comparison on two downstream tasks, where the performance improvements of ReCP are compared with the best of these baseline methods, marked by the asterisk.}
\label{tab:baseline}
\setlength{\tabcolsep}{11pt}
\renewcommand{\arraystretch}{1.2}
\resizebox{0.98\textwidth}{!}{
\begin{tabular}{ c c  c  c  c  c  c  }

\hline
\multirow{2}*{Method} &
\multicolumn{3}{c}{Land Usage Clustering} &
\multicolumn{3}{c}{Region Popularity Prediction} \\
\cline{2-7}
 \multirow{3}*{} & NMI & ARI & F-measure & MAE & RMSE & R$^2$\\
\hline
HDGE &	 0.469 ± 0.01  &  0.095 ± 0.01  &  0.117 ± 0.01  &  334.43 ± 10.17  &  474.94 ± 9.49  &  0.079 ± 0.04  \\
ZE-Mob &	0.437 ± 0.02&	0.071 ± 0.01 &	 0.097 ± 0.01&	 282.42 ± 13.71&	 418.02 ± 12.69&	 0.286 ± 0.04\\
MV-PN &	 0.407 ± 0.01&	 0.036 ± 0.01&	 0.070 ± 0.01&	 291.17 ± 16.54&	 435.23 ± 16.52&	 0.226 ± 0.06\\
CGAL &	 0.414 ± 0.08&	 0.059 ± 0.06&	 0.091 ± 0.06&	 351.10 ± 51.20&	 486.96 ± 52.58& 0.021 ± 0.20	 \\
MVURE &	0.735 ± 0.01& 0.400 ± 0.02& 0.415 ± 0.02& 236.25 ± 7.86& 347.01 ± 11.70& 0.508 ± 0.03\\
MGFN &	   0.748 ± 0.01& 0.424 ± 0.03& 0.437 ± 0.03& 240.37 ± 11.99& 354.24 ± 17.14& 0.487 ± 0.05\\
ReMVC &	 0.761* ± 0.02& 0.455* ± 0.04& 0.462* ± 0.04& 283.02 ± 18.03& 406.25 ± 18.00& 0.325 ± 0.06\\
HREP &	 0.757 ± 0.01& 0.448 ± 0.03& 0.457 ± 0.03& 217.52* ± 10.98& 318.41* ± 14.54& 0.585* ± 0.04
\\
\hline 
\textbf{ReCP} & \textbf{0.780 ± 0.01} & \textbf{0.483 ± 0.01} & \textbf{0.499 ± 0.02} & \textbf{195.16 ± 18.70} & \textbf{291.19 ± 20.04} & \textbf{0.652 ± 0.05} \\
\textbf{Improvements}  &  \textbf{2.50\%} & \textbf{6.15\%}   & \textbf{8.01\%}  & \textbf{10.28\%}  &  \textbf{8.55\%} & \textbf{11.45\%}    \\
\hline
\end{tabular}}
\end{table*}

\textbf{Model Parameters}.
In our experiments, the dimension of region representations is set to 96. In the intra-view reconstruction module, we set the number of layers at 3 and the hidden size at 128 for the encoder $E^{(v)}$ and decoder $D^{(v)}$; in the intra-view contrastive learning module, following the settings in \cite{zhang2022region}, we set the number of positive samples for region attribute and human mobility data at 3 and 4, and the parameter $\mu$ controlling the balance between different views at 0.0001. In the inter-view dual prediction module, we set the number of layers at 3 and the hidden size at 96 for $F^{(a)}$ and  $F^{(m)}$; in the inter-view contrastive learning module, we set the parameter $\alpha$ at 9. We set the hyper-parameters $\lambda_1$ and $\lambda_2$ in the final objective loss at 1. Note that the optimal model parameters are selected using grid search with a small but adaptive step size. To optimize our model, we adopt Adam and initialize the learning rate at 0.01 with a linear decay. 

\textbf{Baselines}.
We compare the performance of ReCP with several state-of-the-art region embedding methods. 
\begin{itemize}
\item  \textbf{HDGE.} \cite{wang2017region} constructs flow graphs and spatial graphs using taxi data and learns region representations with graph embedding methods. 
\item \textbf{ZE-Mob.}  \cite{yao2018representing} models co-occurrence patterns between regions from mobility data to learn region representations. 
\item \textbf{MV-PN.} \cite{fu2019efficient} models both inter-region and intra-region information to construct multi-view POI-POI networks within each region. 
\item \textbf{CGAL.} \cite{zhang2019unifying} extends MV-PN and incorporates the spatial structure and spatial autocorrelation among regions to learn region representations.
\item \textbf{MVURE.} \cite{zhang2021multi} learns region representations by cross-view information sharing and multi-view fusion with human mobility and region attributes.
\item \textbf{MGFN.} \cite{wu2022multi} designs multi-level cross-attention mechanisms to extract region representations from multiple mobility patterns.
\item \textbf{ReMVC.} \cite{zhang2022region} learns region representations through both intra-view and inter-view contrastive learning modules. 
\item \textbf{HREP.} \cite{zhou2023heterogeneous} constructs heterogeneous graphs and uses relation-aware graph embedding to learn region representations.
\end{itemize}

\subsection{Land Usage Clustering}
We use the district division by the community boards \cite{berg2007new} as ground truth and divide the Manhattan borough into 29 districts, following the settings in \cite{zhang2022region}. We cluster regions into groups by $k$-means clustering ($k=29$), using region representations as inputs. The regions with the same land usage type are expected to be assigned to the same cluster. The experimental results are evaluated using three metrics: Normalized Mutual Information (NMI), Adjusted Rand Index (ARI), and F-measure following \cite{yao2018representing, zhang2021multi}. We assess all the methods using the same dataset and conduct 10 runs to report the mean value with the standard deviation in Table \ref{tab:baseline}. From the results, we observe that:
\begin{itemize}
    \item HDGE and ZE-Mob exhibit relatively inferior performance as they merely model co-occurrence patterns using human mobility data. MGFN demonstrates better performance than HDGE and ZE-Mob, as it designs a deep model based on cross-attention mechanisms to capture complex mobility patterns from spatial-temporal human mobility data. 

    \item The methods that model multi-view information generally achieve satisfactory results, validating the importance of effectively integrating multi-view information for region embedding. Specifically, MV-PN and CGAL exhibit poor performance as they simply combine region representations from two views, lacking the deep interaction between views; MVURE and HREP design attention-based mechanisms to fuse the multi-view information, consequently yielding superior performance; ReMVC adopts contrastive learning to model intra-view and inter-view information and also obtains good results.

    \item The proposed ReCP outperforms all these baselines, as it explores the consistency across different views in region embedding. Compared with ReMVC, ReCP achieves average improvements of 2.50\%, 6.15\%, and 8.01\% in terms of NMI, ARI, and F-measure, respectively.  Moreover, the results of the superiority paired t-test indicate that the improvement of ReCP over the baselines is statistically significant, with a $p$-value less than 0.01.
\end{itemize}


\subsection{Region Popularity Prediction}
Another commonly-compared downstream task to evaluate the region representations is popularity prediction, where we aggregate the check-in counts within each region as the ground truth of popularity following \cite{yang2014modeling, zhang2022region}. We take region representations as input and train the Lasso regression model. The evaluation results including Mean Absolute Error (MAE), Root Mean Square Error (RMSE), and Coefficient of Determination (R$^2$) are obtained by 5-fold cross-validation, as reported in Table \ref{tab:baseline}. From the results, we observe that the multi-view fusion methods including MVURE and HREP achieve decent performance, which further validates the necessity of integrating multi-view information in region embedding. ReCP performs the best among all methods, e.g., compared to HREP, ReCP achieves average improvements of 10.28\%, 8.55\%, and 11.45\% in terms of MAE, RMSE, and R$^2$. These results indicate that it is an effective way to learn better region representations by utilizing the new pipeline following the consistency learning paradigm.


\subsection{Ablation Study and Parameter Analysis}
\subsubsection{Ablation study} 
\begin{figure}[!t]
\captionsetup[subfigure]{labelformat=simple}
    \centering
    \subfloat[Land Usage Clustering]{
    \label{fig:dataSparsityJP}
    \includegraphics[width=0.155\textwidth]{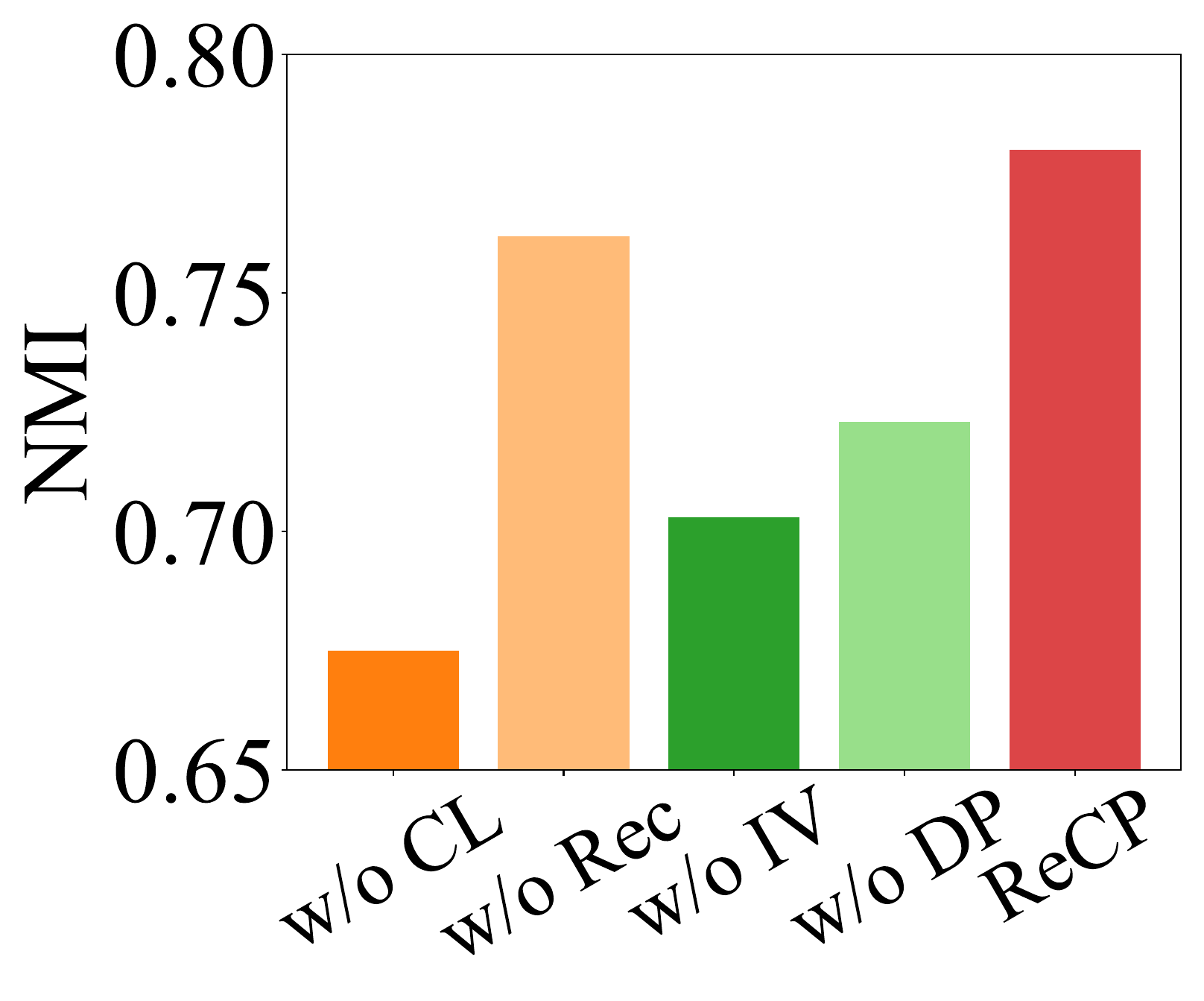}
    \hspace{-1mm}
    \label{fig:dataSparsityUS}
    \includegraphics[width=0.155\textwidth]{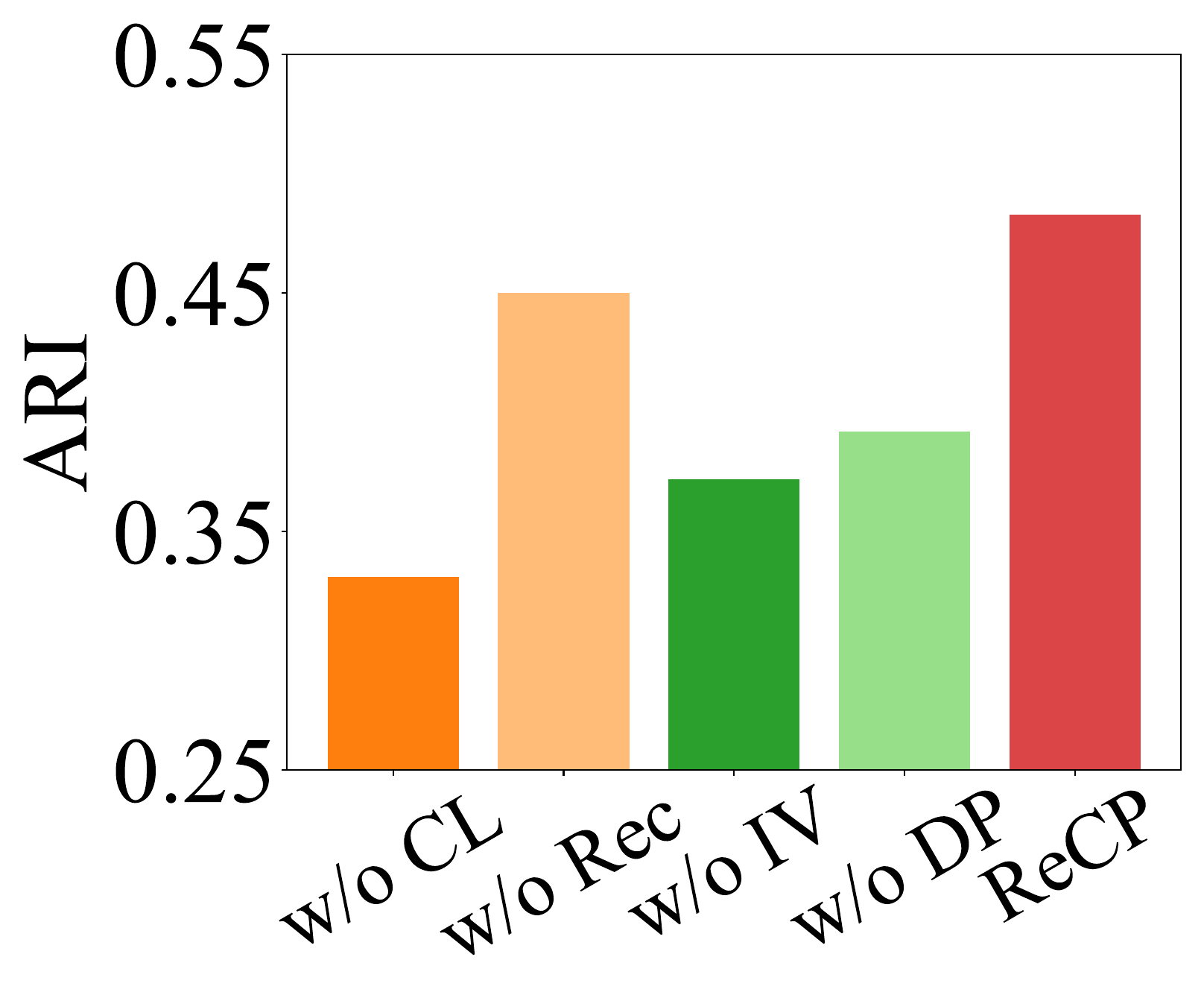}
    \hspace{-1mm}
    \label{fig:alpha_JP}
    \includegraphics[width=0.155\textwidth]{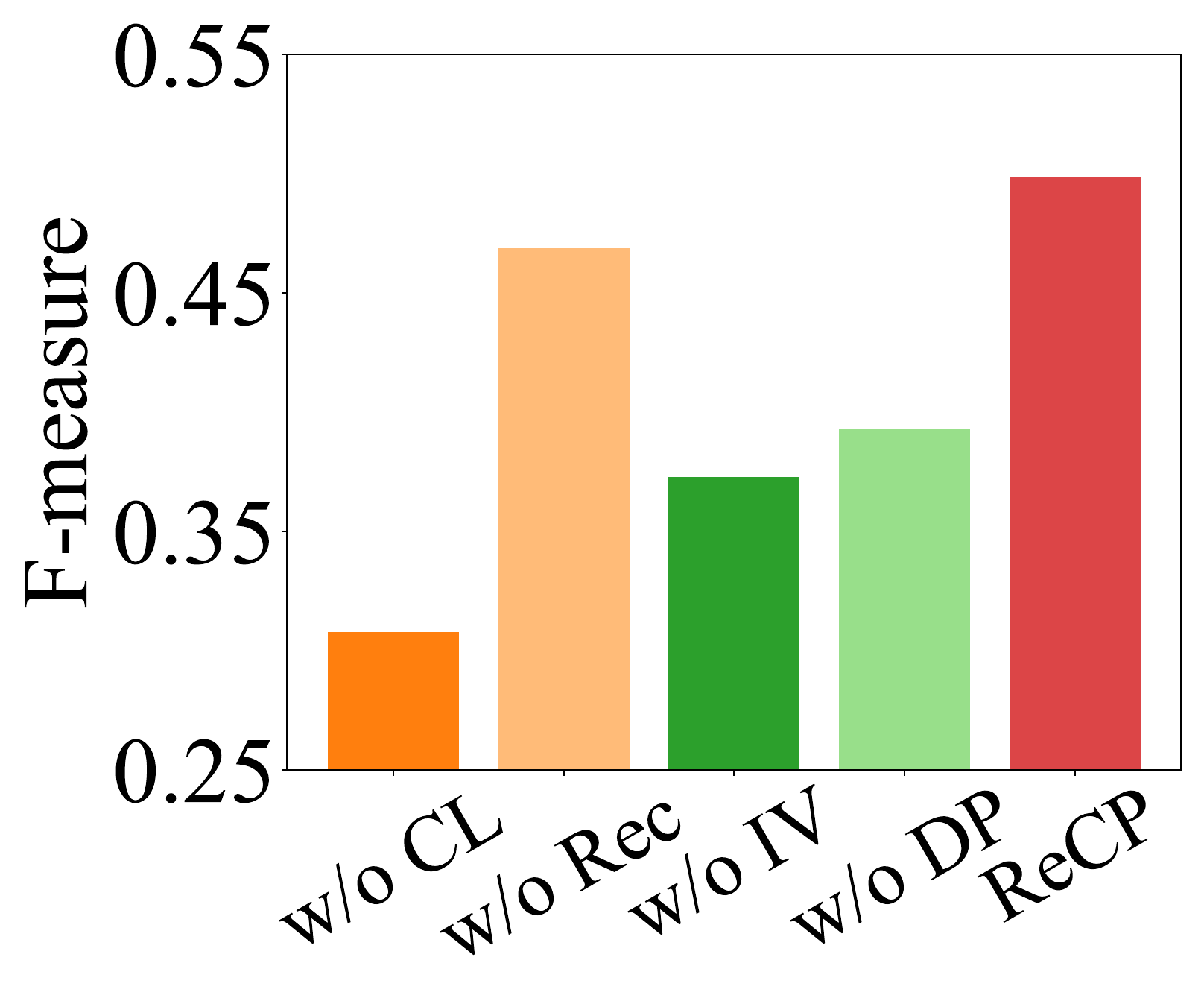}}
    
    
  \subfloat[Region Popularity Prediction]{
    \label{fig:alpha_US}
    \includegraphics[width=0.155\textwidth]{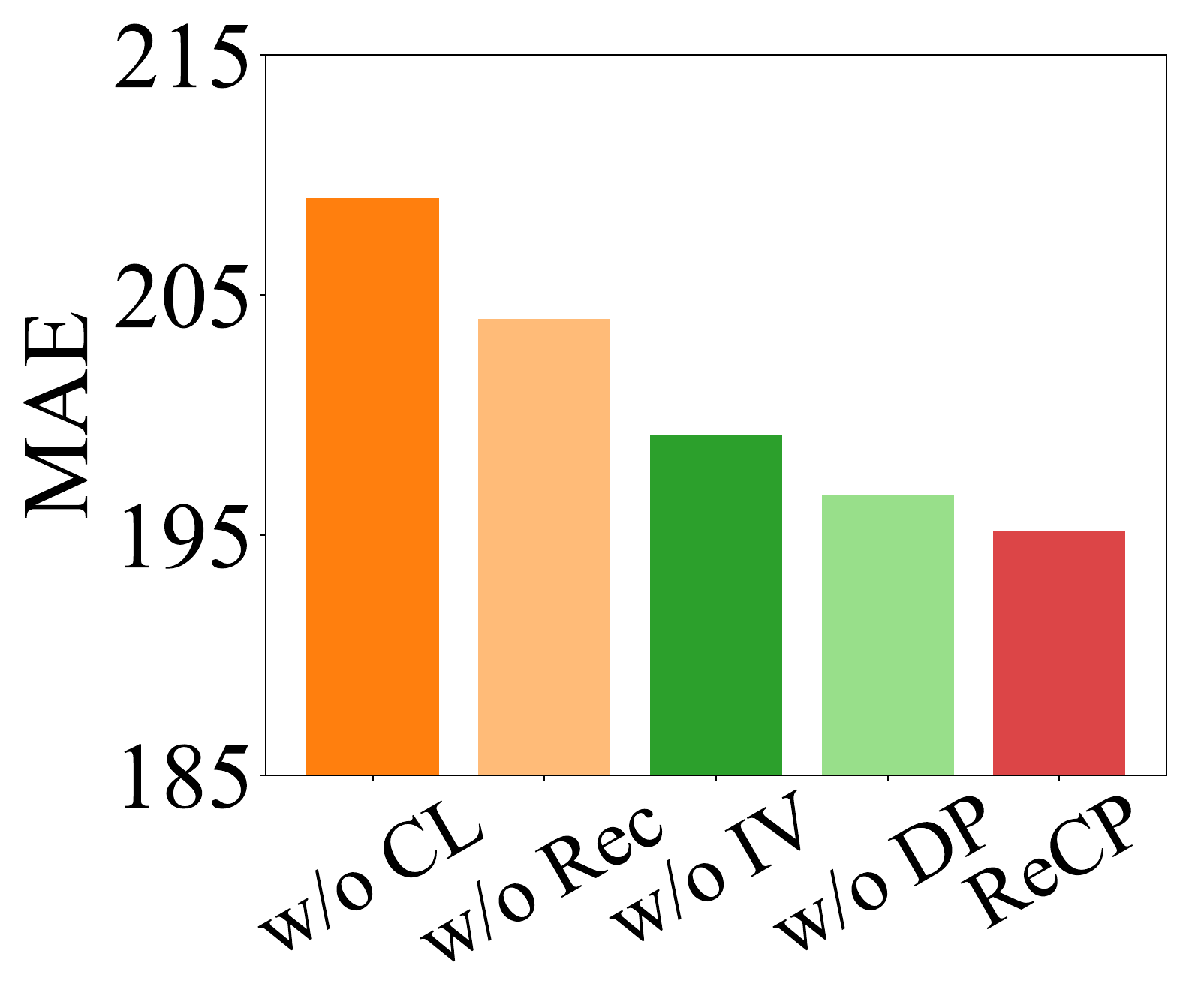}
    \hspace{-1mm}
    \label{fig:alpha_US}
    \includegraphics[width=0.155\textwidth]{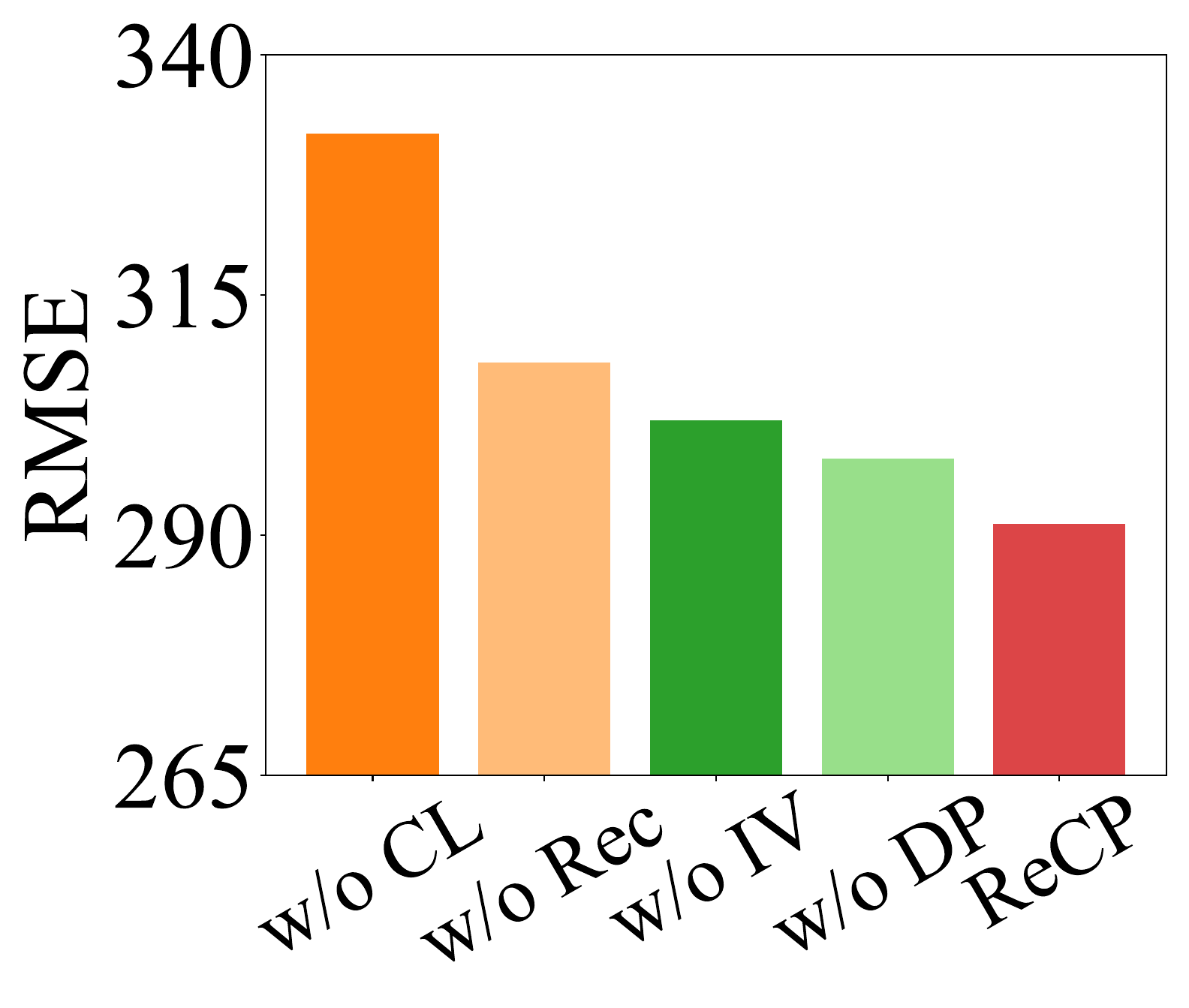}
    \hspace{-1mm}
    \label{fig:alpha_US}
    \includegraphics[width=0.155\textwidth]{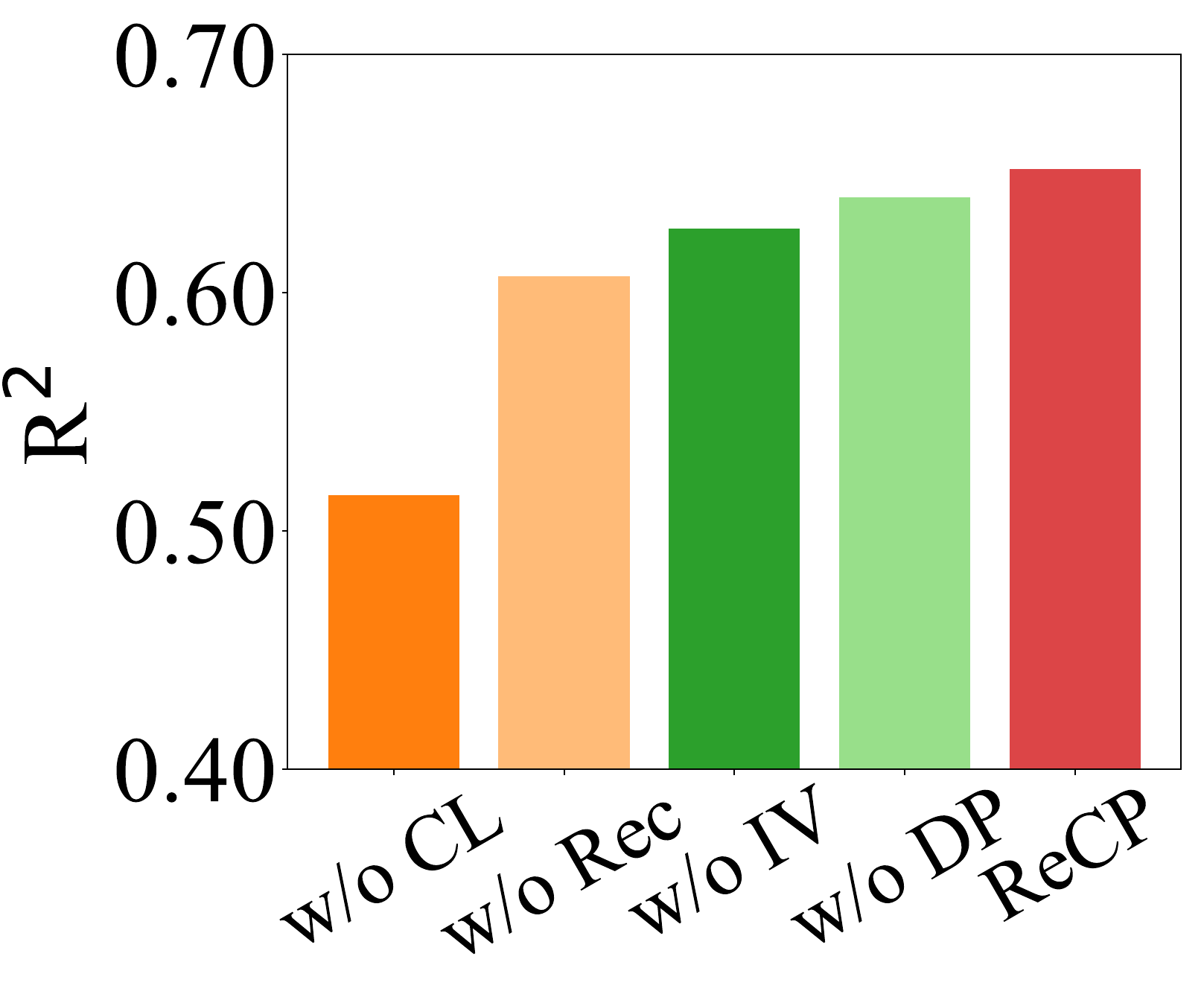}}
  
    \caption{Performance comparison of different modules.}
    \label{fig:ablation}
\end{figure}

We design four variants to explore how each module of ReCP affects the clustering and regression performance:
\begin{itemize}
 \item ReCP w/o CL: we remove the intra-view contrastive learning loss.
 \item  ReCP w/o Rec: we remove the intra-view reconstruction loss and only use the encoder to extract features.
 \item  ReCP w/o IV: we remove the inter-view learning module and directly concatenate region representations from the two views without the constraint of consistency learning.
 \item  ReCP w/o DP: we remove the inter-view dual prediction loss. 
\end{itemize}
From the results in Figure~\ref{fig:ablation}, we observe that: 

1) ReCP w/o CL achieves the lowest performance in both tasks, indicating that the intra-view contrastive learning loss is a crucial component in our model for learning view-specific feature representations of regions.

2) ReCP w/o Rec achieves worse performance than ReCP, supporting the aforementioned claim that the intra-view reconstruction loss could help prevent the model from converging to a trivial solution. 

3) ReCP demonstrates an improvement of 29.84\% (in terms of ARI) and 4.00\% (in terms of R$^2$) when compared to ReCP w/o IV. This finding suggests that the proposed inter-view learning module effectively leverages the multi-view information and highlights the importance of consistency learning across different views.

4) ReCP w/o DP outperforms ReCP w/o IV but performs worse than ReCP, indicating that both the inter-view contrastive learning loss (which maximizes the mutual information between views) and the inter-view dual prediction loss (which minimizes the conditional entropy across them) are important for learning multi-view region representations.


\subsubsection{Parameter sensitivity}

\begin{figure}[!t]
\captionsetup[subfigure]{labelformat=simple}
    \centering
    \subfloat[Land Usage Clustering]{
    \label{fig:dataSparsityJP}
    \includegraphics[width=0.15\textwidth]{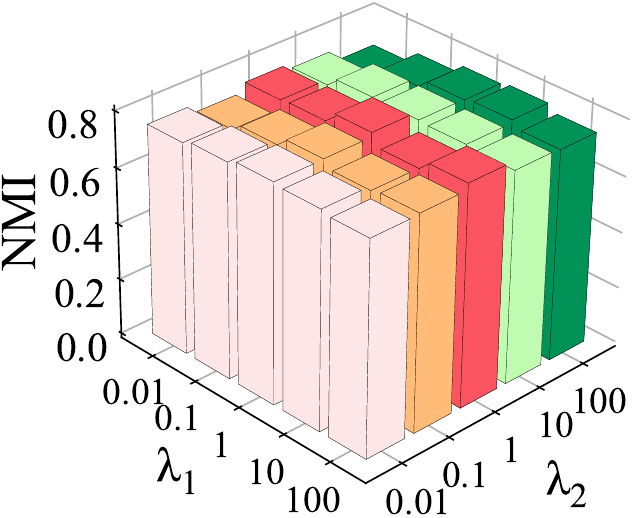}
    \hspace{0.05mm}
    \label{fig:dataSparsityUS}
    \includegraphics[width=0.15\textwidth]{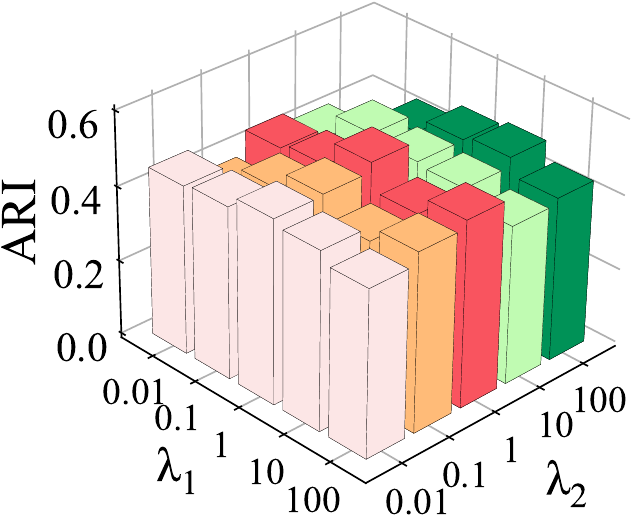}
    \hspace{0.05mm}
    \label{fig:alpha_JP}
    \includegraphics[width=0.15\textwidth]{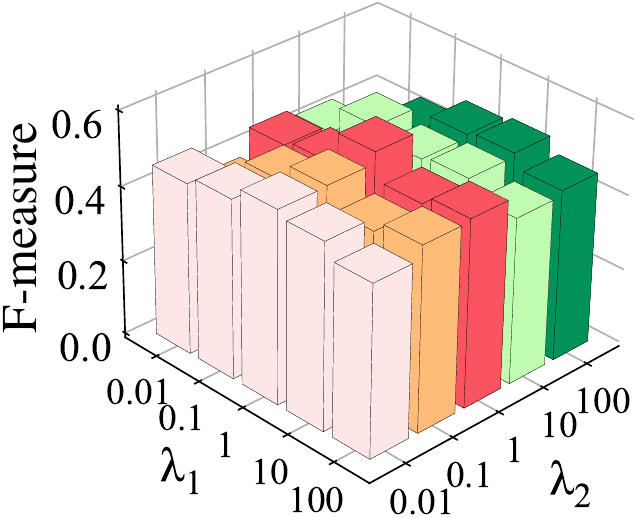}
    }

    \subfloat[Region Popularity Prediction]{
    \label{fig:alpha_US}
    \includegraphics[width=0.15\textwidth]{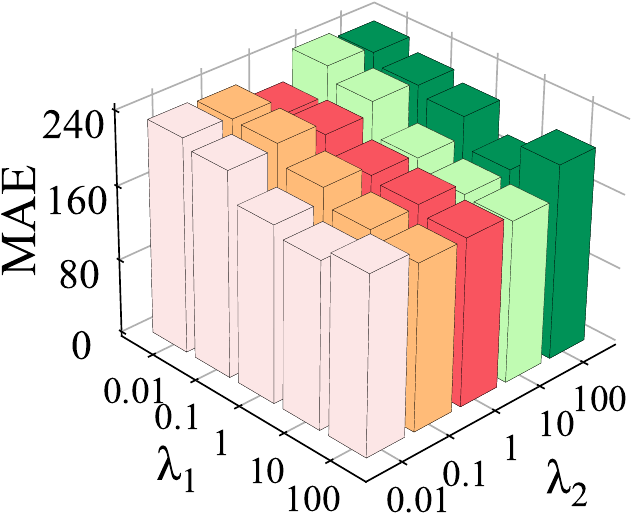}
    \hspace{0.05mm}
    \label{fig:alpha_US}
    \includegraphics[width=0.15\textwidth]{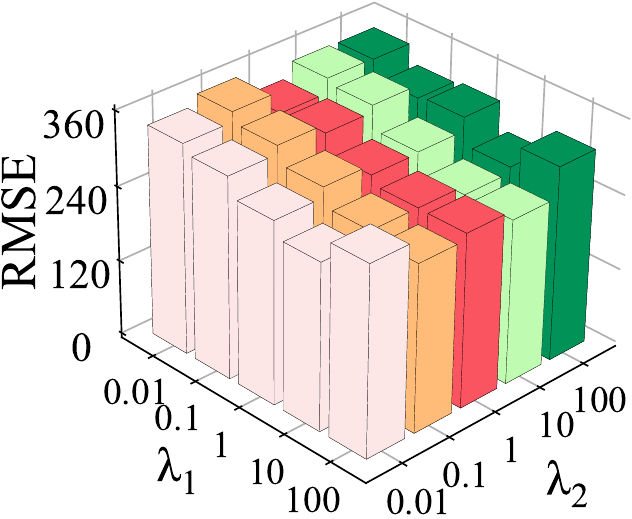}
    \hspace{0.05mm}
    \label{fig:alpha_US}
    \includegraphics[width=0.15\textwidth]{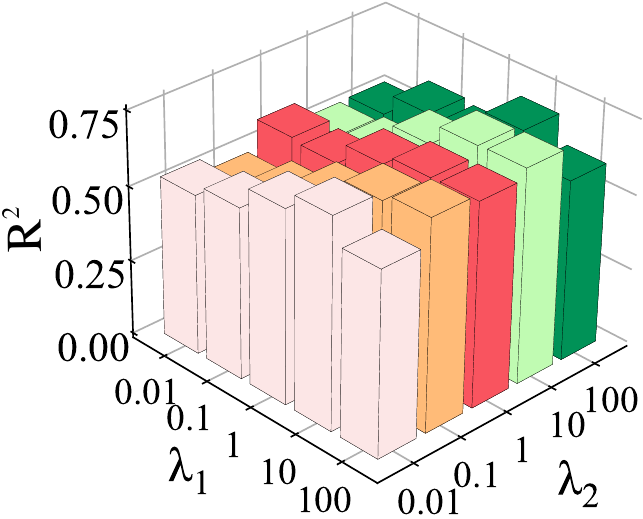}
    }
    \caption{Parameter analysis on both downstream tasks.}
    \label{fig:modelanalysisofncp}
\end{figure}

The parameters $\lambda_1$ and $\lambda_2$ govern the weighting of various losses. We vary their values within the range of $\{0.01, 0.1, 1, 10, 100\}$ to assess the impact of $\lambda_1$ and $\lambda_2$ on the model performance. As depicted in Figure~\ref{fig:modelanalysisofncp}, ReCP achieves satisfactory performance when we set both $\lambda_1$ and $\lambda_2$ at 1.

\section{Related Work}\label{relatedwork}



Traditional methods for region embedding typically utilize human mobility data to analyze the transition patterns between urban regions. These methods are often based on the word2vec framework and learn the latent representations of regions \cite{wang2017region, yao2018representing}. In a similar vein, Wu et al. \cite{wu2022multi} incorporate mobility graphs with spatio-temporal similarity as mobility patterns, and propose multi-level cross-attention mechanisms to extract comprehensive region representations from these patterns. Additionally, some studies focus on leveraging the inherent attributes of regions to learn latent representations. For instance, Zhang et al. \cite{zhang2019unifying} construct multiple spatial graphs to represent the geographic structure of regions. By transforming the region embedding problem into a graph embedding problem, they primarily capture the spatial structure within regions and the spatial autocorrelation between regions. Another approach, proposed by Wang et al. \cite{wang2020urban2vec}, involves mining street views and textual information of POIs within regions to learn representations.


Moreover, there have been studies that learn region representations by incorporating both attribute features within regions and mobility patterns between regions. For instance, Fu et al. \cite{fu2019efficient} propose an autoencoder framework that effectively captures inter-region correlations and intra-region structural information during the process of region embedding. Zhang et al. \cite{zhang2021multi} model multi-view region correlations by leveraging human mobility data and inherent region attributes, and employ a graph attention mechanism to acquire region representations from each view of the established correlations. Zhou et al. \cite{zhou2023heterogeneous} learn relation-specific region representations from various types of relations in a heterogeneous graph constructed using human mobility, POI data, and geographic neighbors of regions. They devise an attention-based fusion technique to integrate shared information among different types of correlations. Additionally, Zhang et al. \cite{zhang2022region} introduce a multi-view region embedding model based on contrastive learning, which incorporates an intra-view contrastive learning module to discern distinct representations and an inter-view contrastive learning module to facilitate the transfer of knowledge across multiple views.


\section{Conclusion}
In this paper, we form a new pipeline based on the consistency learning paradigm for multi-view region embedding. Under the hood, we propose a multi-view Contrastive Prediction model for urban Region embedding (ReCP) by exploring the consistency across two views, leveraging both POI and human mobility data. The ReCP model consists of two modules: an intra-view learning module that utilizes contrastive learning and feature reconstruction which learn region representations specific to each view, and an inter-view learning module utilizing a contrastive prediction learning scheme that enhances the consistency between two views. To evaluate the effectiveness of our proposed model, we conduct comprehensive experiments on two downstream tasks: land use clustering and region popularity prediction. The experimental results demonstrate that the proposed ReCP model outperforms state-of-the-art embedding methods, proving that retaining consistency across views is pivotal for effective region embedding.

\clearpage

\bibliography{ref}

\end{document}